# ARC-2: Continuous Compliance Through Delta Management

Tord Johan Eide, Einar Holt
Wybe Labs Inc.

## Abstract

ARC [1] introduced a three-layer governance architecture for autonomous systems consisting of model safety validation, cognitive certification (CRAB), and operational authorization (LEA). While ARC defines the conditions required for deployment, it does not specify how compliance should be maintained once systems enter operational environments. This paper addresses that gap.

We introduce Delta Management, a framework for continuous post-deployment governance in which the certified baseline established during authorization serves as a permanent reference state — more precisely, a certified region rather than a point. Operational compliance is defined as the measurable distance between the current system state and this certified baseline region. Governance decisions are therefore driven not by static certification status, but by continuous measurement of behavioral deviation over time.

Building directly upon the ARC architecture, we introduce four extensions. First, CRAB evaluation is anchored to the worst verified behavioral deviation, with single-observation outliers quarantined pending confirmation, ensuring that demonstrated failure modes define the operational safety floor without creating a monotonically non-increasing authority that degenerates over time. Second, we replace a single AND-gated spiral trigger with a layered trigger taxonomy that distinguishes hard safety invariants, critical single-dimension overrides, confirmed multi-system spirals, and risk-weighted composite scores — each coupled to appropriately scaled governance responses. Third, we formalize safe transition as a graduated protocol that explicitly separates governance authority withdrawal from physical stop, aligned with IEC 60204-1, ISO 13850, SAE J3016, and EU AI Act Article 14(4)(e). Fourth, we introduce environmental delta as a distinct baseline validity condition: when the operational environment drifts beyond the certified Operational Design Domain (ODD), the response is recertification or ODD revision — not authority contraction — because the baseline itself has become an invalid compliance reference.

Together, these contributions transform ARC from a deployment-time certification architecture into a continuous governance framework capable of monitoring, evaluating, and enforcing compliance throughout the operational lifecycle of autonomous systems. The framework is designed for explicit positioning against EU AI Act Article 72(2), which requires providers of high-risk AI systems to evaluate their continuous compliance throughout the system lifecycle.

## 1. Introduction

Autonomous robotic systems are deployed in environments that differ materially from the controlled conditions under which they are developed and tested. A warehouse robot operates in a facility whose layout, occupancy, and task distribution change over time. A healthcare robot encounters patients, staff, and environmental configurations it has never seen in testing. A logistics system executes plans derived from models that were accurate at training time but may not remain so.

ARC [1] proposed that autonomous robotic systems require governance across three distinct domains: the safety properties of the underlying model, the cognitive capabilities demonstrated under certification, and the operational scope within which the system is authorized to act. The architecture was motivated by the inadequacy of single-point validation — the assumption that a system safe in a laboratory remains safe in a non-scripted operational environment.

That critique holds. But it leaves a gap. ARC defined the structure of the certification stack. It did not define the mechanism by which that stack responds to what happens after deployment.

Deployed autonomous systems are not static. Models drift. Environments change. Edge cases accumulate. A governance architecture that only asks "was this system certified?" cannot answer the question that matters in operation: "is this system still compliant with its certified baseline?" The distinction between these questions is the difference between a single-point safety check and a continuous governance framework.

This paper proposes delta management as the answer. The delta — the measurable distance between a certified baseline region and current operational state — is the primary compliance signal. The governance rules that follow are natural consequences of a single principle: the authorized operational envelope is defined at certification, and everything that follows is the management of distance from that definition.

The paper is organized as follows. Section 2 summarizes the ARC architecture this paper extends. Section 3 introduces the delta management principle and its four dimensions. Sections 4 through 6 address CRAB governance, behavioral spiral detection, and asymmetric authorization. Section 7 addresses implementation implications. Section 8 positions the contribution against related work and regulatory context. Section 9 concludes.

## 2. Background: The ARC Stack

ARC [1] introduced a three-layer governance architecture argued on necessity grounds: each layer addresses governance questions the others cannot resolve. No claim of sufficiency is made — post-market surveillance, supply-chain hardware integrity, and cybersecurity governance may be required in parallel, depending on deployment context and applicable regulatory requirements.

### 2.1 Model Safety Validation

The first layer addresses the safety properties of the underlying AI model before cognitive evaluation begins. Model safety validation establishes whether the system's foundational behavior — its outputs under a defined range of inputs — satisfies minimum safety requirements. This layer operates independently of task performance: a model may perform well on average while retaining latent failure modes that become active only in specific edge conditions. Model safety validation is the prerequisite for the cognitive evaluation layer that follows.

### 2.2 Cognitive Certification: CRAB

The second layer is the Cognitive Robotics Accreditation Benchmarks (CRAB), a structured evaluation of the cognitive capabilities demonstrated by the autonomous system under controlled conditions. CRAB evaluates capabilities across eight dimensions in two tiers. Tier 1 (core cognitive): situational awareness (SA), planning and execution (UP), human interaction (HI), self-correction (SC), generalization (CG), and energy and efficiency (CE). Tier 2 (trust and governance, double-weighted in domain suite evaluations): safety and constraint adherence (SCA) and decision transparency (DT). Evaluation produces a per-dimension capability profile on a 0–1000 scale; the profile, not any aggregate score, is the primary governance evidence. Domain suites (CRAB-H for healthcare, CRAB-A for administrative, CRAB-I for industrial) may apply mandatory minimum floors to Tier 2 dimensions and, for high-risk contact classes, supplementary floors to Tier 1 dimensions. The CRAB evaluation produces a certified capability profile rather than a single aggregate score, preserving the dimensional structure that matters for task-specific governance.

The necessity argument for CRAB is straightforward: model safety validation alone cannot establish whether a system possesses the specific cognitive capabilities required for safe autonomous operation. A model that passes safety validation may still lack adequate situational awareness, fail to recognize when human intervention is required, or be unable to explain its decisions in ways that support accountability. These are governance questions that model-level evaluation cannot resolve.

### 2.3 Operational Authorization: LEA

The third layer is the Layered Envelope Authorization (LEA), which determines the operational scope within which a certified system may act. LEA authorization specifies the tasks, environments, interaction modes, and escalation conditions under which the system is permitted to operate. Authorization is granted not to the system in isolation but to the relationship between demonstrated capability and the requirements of a specific operational context.

The necessity argument for LEA is equally direct: CRAB certification establishes what a system can do; it does not establish what the system is permitted to do in a particular environment. Demonstrated capability does not imply permission to exercise that capability in every operational context. LEA maintains this separation, ensuring that authorization decisions remain independent of — and conditional upon — certification outcomes.

### 2.4 The Gap ARC-2 Addresses

ARC defines the conditions required for deployment. What it does not define is the mechanism by which the governance stack responds to post-deployment operational reality. The certified baseline established at authorization is the foundation of ARC-2's delta management framework. Delta Management does not replace the three ARC layers; it animates them continuously throughout the operational lifecycle.

## 3. The Delta Management Principle

### 3.1 Compliance as Distance

Let B denote the certified baseline — not a single point but a certified region encompassing the full system state at the point of LEA authorization, including the environmental distribution under which certification testing was conducted. Let S(t) denote the observed system state at time t. The compliance signal is:

$$\Delta(t) = d(S(t), B)$$

where $d(S(t), B) = \inf_{b \in B} \|S(t) - b\|$ is the distance-to-set function, equal to zero when S(t) lies within the certified region B and positive when S(t) lies outside it. Compliance exists when $\Delta(t) = 0$ (the system remains within the certified region) or when $\Delta(t)$ remains within an authorized tolerance margin. Exceeding that margin constitutes a governance event requiring a predefined response. Compliance exists when $\Delta(t)$ remains within an authorized tolerance envelope. Exceeding that envelope does not necessarily imply system failure; rather, it constitutes a governance event requiring a predefined response.

This framing has three immediate consequences. First, governance becomes continuous — there is no moment in the operational lifecycle at which the compliance question is not live. Second, the mechanism of failure becomes precise — drift is detectable before it becomes catastrophic. Third, the burden of proof shifts — a system must not only perform; it must demonstrate that its performance remains within certified parameters.

It is important to note that Δ(t) is not directly observable as a single scalar quantity. In practice, Δ(t) constitutes a vector of observable proxies rather than a computable scalar. The governance architecture must specify, as part of each LEA authorization, which proxies constitute the compliance signal for each delta dimension, what measurement cadence applies, and under what conditions proxy-level readings trigger a full cross-layer evaluation. Treating Δ(t) as a single computable scalar overstates the precision of the current art; the framework is most honestly characterized as a structured inference problem operating over a monitored proxy space. Suitable divergence measures — including the Maximum Mean Discrepancy (MMD) and Wasserstein distance — may be appropriate for specific delta dimensions where distributional comparison is required [2].

### 3.2 Four Dimensions of Delta

Delta is not scalar. A fully specified compliance signal requires measurement across four dimensions, corresponding to ARC's three layers plus the environmental validity condition introduced in Section 3.3:

(1) Model delta — deviation in safety-relevant model behavior from validated baseline, assessed through behavioral fingerprints, statistical output signatures, version hashes, and update attestation records. For systems with continual learning, LoRA adapters, or retrieval-augmented generation, version hashes are unstable by design; the appropriate formulation is delta against a pre-approved change envelope, analogous to the FDA's Predetermined Change Control Plan framework [3].

(2) Cognitive delta — deviation in CRAB dimension scores from certified benchmark profile, assessed through continuously monitored behavioral proxies calibrated against the certified baseline. Full CRAB re-evaluation is a triggered governance response, not a routine monitoring event.

(3) Operational delta — deviation in task performance, environmental fit, and edge-case handling from authorized envelope parameters, captured through task performance telemetry, sensor logs, and actuator data.

(4) Environmental delta — deviation of the operational environment from the environmental distribution under which certification was conducted. This dimension is categorically distinct from the other three: it measures the world, not the system. See Section 3.3.

A governance architecture that monitors only one or two dimensions will miss correlated failure across layers — which is the most dangerous failure mode in deployed autonomous systems.

### 3.3 Environmental Delta as Baseline Validity Condition

The delta management framework depends on the stability of B — the certified baseline region — as a reference against which operational compliance is continuously measured. Environmental delta introduces a condition that the other three dimensions cannot capture: whether B remains a valid compliance reference at all.

A system may exhibit no internal change — model weights unchanged, CRAB scores stable, task performance nominal — while its operational environment drifts materially from the conditions under which it was certified. In such cases, the system's behavioral delta relative to its certified baseline may grow not because the system has degraded, but because the baseline is no longer a valid reference for the environment in which the system is operating. This phenomenon is well-established in the dataset shift and covariate shift literature [4] and in post-market monitoring frameworks for medical AI systems [5].

Environmental delta is therefore a baseline validity condition with a distinct governance response. When model, cognitive, or operational delta exceeds defined thresholds, the response is authority contraction, re-evaluation, suspension, or revocation — responses directed at the system. When environmental delta exceeds a defined threshold — measured through distributional distance metrics applied to operational sensor data relative to the certification environment — the required response is recertification under the new environmental conditions or revision of the certified ODD, because the baseline itself has become an inadequate compliance reference.

ARC-2 proposes that LEA authorization explicitly include a specification of the environmental envelope within which the certified baseline is valid, and that monitoring of environmental distribution shift be a mandatory component of the continuous governance stack, triggering a recertification review rather than authority contraction when thresholds are exceeded.

## 4. CRAB Governance Based on Worst-Case Behaviour

### 4.1 The Inadequacy of Mean-Based Scoring

A system that scores 0.85/1.0 on average across CRAB dimensions but has demonstrated a single-instance collapse to 0.20/1.0 in a specific environmental context presents a categorically different risk profile than a system that scores 0.60/1.0 consistently. Mean-based performance metrics obscure demonstrated failure modes and may substantially underestimate operational risk. All CRAB scores in this framework use per-dimension vectors normalized to 0–1, preserving dimensional structure for task-specific governance.

### 4.2 The Outlier Floor Principle

ARC-2 proposes: the authorized CRAB floor is anchored to worst observed delta, not mean performance. The rationale is precisely the ARC-1 argument applied inward: just as single-point pre-deployment validation is a false floor for safety, single-period average CRAB scoring is a false floor for ongoing compliance.

However, the naive running-minimum formulation contains a structural defect: a running minimum has breakdown point zero — at any positive contamination rate (sensor faults, miscalibrated evaluators, adversarial inputs), the probability of producing a spurious floor reading approaches 1 over time. Without a reset pathway, governance authority becomes monotonically non-increasing and degenerates toward zero for any long-deployed system. This is a structural guarantee of failure, not merely an incentive problem.

The outlier floor principle addresses this through three mechanisms:

(a) Quarantine before binding: a single CRAB reading below a defined threshold is placed in quarantine and does not immediately bind the governance floor. A floor observation binds only upon confirmation — either through a repeated observation meeting the threshold, or through validated provenance establishing that the observation was not the product of measurement error. This preserves the worst-case anchoring rationale while preventing individual sensor faults from permanently degrading the operational envelope.

(b) Separation of raw logs from governance state: append-only, tamper-evident raw telemetry logs (required under AI Act Articles 12 and 72 [6, 7]) must be strictly separated from the revisable governance state used to compute current compliance. Raw logs are immutable by design; governance state reflects the current best estimate of authorized operational parameters and may be updated through the recertification pathway described below.

(c) Explicit rebaseline pathway: following recertification — whether triggered by confirmed outlier observations, environmental drift, or model update — the certified baseline region B is formally updated. Historical observations that contributed to the previous floor do not carry over to the new certification governance floor. However, any confirmed failure mode — a quarantine-confirmed outlier observation that triggered the rebaseline — must be incorporated into the recertification test set as a mandatory scenario. The raw logs remain intact for audit purposes; the governance floor reflects the new certified baseline; but the recertification evaluation must demonstrate that known failure modes have been addressed, not merely that the system performs acceptably on a fresh scenario set.

### 4.3 Reliability-Gated Assurance Allocation

CRAB-H [25] introduces a reliability threshold mechanism for cognitive certification dimensions in healthcare autonomous systems. The same principle applies to any CRAB-certified domain. Dimensions that cannot be evaluated with inter-rater reliability meeting defined thresholds — weighted $\kappa \geq 0.70$ for standard dimensions, weighted $\kappa \geq 0.80$ for safety-critical and dignity-critical dimensions, applied to the lower bound of the 95% confidence interval — are reclassified from CRAB certification to LEA operational requirements.

This reliability-gated assurance allocation preserves the governance architecture's integrity: unreliable dimensions are not ignored; they are governed at the operational authorization layer rather than the certification layer. The delta management framework described in this paper applies to both CRAB-certified dimensions and LEA-governed dimensions, with proxy monitoring designed to the reliability characteristics of each dimension.

The mechanism is illustrated in TRP-H-001 (wheelchair transfer) : a dignity preservation dimension — suppose a hypothetical pilot evaluation yields weighted $\kappa = 0.58$ (95% CI: 0.41–0.75) — falling below the $\geq 0.80$ safety-critical threshold. The dimension was rerouted to LEA as an operative requirement — human supervision of dignity aspects throughout every transfer — rather than discarded or inflated with false precision.[25]

### 4.4 Triggering Re-Evaluation

Any confirmed CRAB reading below a defined threshold — or any confirmed reading representing a statistically significant deviation from the certified baseline — triggers mandatory re-evaluation. Re-evaluation is not discretionary; it is a governance event with a defined latency requirement specified as part of the LEA authorization.

Continuous CRAB monitoring operates through behavioral proxies, not continuous full CRAB test-battery execution. Proxy streaming must be low-latency and always-on; full evaluation requires access to the certification test environment or its equivalent and carries materially higher computational overhead. The LEA authorization must specify both the proxy set and the threshold conditions that constitute a re-evaluation trigger.

## 5. Behavioral Spirals: Detection, Response, and Staged Revocation

### 5.1 A Layered Trigger Taxonomy

A natural formulation defines a behavioral spiral as requiring all three diagnostic dimensions — deviation magnitude, deviation velocity, and cross-subsystem correlation — to simultaneously exceed defined thresholds. This AND-gated rule preserves a precise definition of correlated multi-system degradation, but it creates a structural problem: a rapid, severe single-subsystem failure does not constitute a spiral under this definition and therefore triggers no governance response. Using spiral detection as the sole escalation pathway maximizes dangerous false negatives.

A further problem is categorical: correlation measures coupling structure, not severity. Combining a structural dimension (cross-subsystem correlation) with severity dimensions (magnitude, velocity) in a uniform voting rule is categorically incoherent. At realistic spiral base rates, all voting rules — AND, OR, and 2-of-3 — will produce poor positive predictive value; the correct response is to couple low-PPV triggers to appropriately low-cost responses rather than to change the trigger rule alone [8].

ARC-2 addresses this through a layered trigger taxonomy. The taxonomy preserves the spiral definition as a precise characterization of one failure mode while creating distinct pathways for other failure patterns:

Layer 1 — Hard Safety Invariants (OR semantics): collision detection, force limit breach, and sudden ODD exit — defined as an abrupt, discrete transition outside the authorized operational boundary (e.g., a door opens into an uncleared zone; the system is moved to an unauthorized area) — unconditionally trigger the safe transition protocol described in Section 5.2, regardless of composite scores. These invariants are not compliance thresholds; they are absolute safety conditions that tolerate no governance delay. They correspond to the stop categories defined in IEC 60204-1 [9] and the emergency stop provisions of ISO 13850 [10].

Gradual environmental drift — the progressive distributional shift of operational conditions away from the certification environment — is categorically distinct from sudden ODD exit. Gradual drift is monitored through the environmental delta dimension (Section 3.3) and triggers recertification or ODD revision when thresholds are exceeded, not the safe transition protocol. A sudden ODD exit is a safety event requiring immediate physical response; gradual drift is a baseline validity condition requiring a governance process.

Layer 2 — Critical Single-Dimension Overrides: extreme magnitude in any single CRAB dimension — defined as a per-dimension deviation exceeding a critical threshold — triggers immediate scope contraction and mandatory re-evaluation. This is not spiral detection; it is a separate trigger class for unidimensional catastrophic failure. A robot that loses the ability to recognize human presence in its operational environment requires immediate response regardless of whether other subsystems are nominal.

Layer 3 — Spiral Detection (AND semantics preserved): a behavioral spiral is confirmed when all three dimensions — magnitude, velocity, and cross-subsystem correlation — exceed defined thresholds. Correlation remains the critical discriminator: uncorrelated single-subsystem deviations suggest local faults; correlated multi-subsystem drift suggests systemic model degradation or environment-model mismatch. Confirmed spirals trigger the full staged response protocol.

Layer 4 — Risk-Weighted Composite Score: a continuously computed risk-weighted composite of all monitored dimensions, calibrated per domain using Average Run Length metrics ($ARL_0$), triggers graduated low-cost responses — increased monitoring frequency, reduced operational scope, additional human oversight requirements — before harder responses are considered. Where domain calibration data is unavailable, a provisional fallback may use a 2-of-3 voting rule across magnitude, velocity, and a simplified binary coupling indicator. Correlation must not be used as a direct vote alongside magnitude and velocity — this conflates structural and severity dimensions in precisely the way §5.1 identifies as incoherent. A binary coupling indicator for provisional use is: cross-subsystem coupling is either present (two or more subsystems simultaneously deviating beyond a defined threshold) or absent. This preserves the logical separation between severity (magnitude, velocity) and structure (coupling) while providing a deployable fallback pending domain calibration.

The threshold values defining trigger confirmation require empirical calibration per system type, operational domain, and deployment environment. This calibration problem is non-trivial and must be documented as part of LEA authorization, including the process by which thresholds are set, the conditions under which they are revised, and the evidence standard required to adjust them post-deployment. The authors do not possess calibration data for specific deployment domains and explicitly invite domain practitioners — in healthcare robotics, logistics automation, and other safety-critical settings — to contribute threshold data, false positive and false negative observations, and domain-specific $ARL_0$ estimates. Such contributions would substantially strengthen the empirical basis of the trigger taxonomy.

### 5.2 Safe Transition Protocol

A critical deficiency in many governance formulations is the implicit assumption that governance authority withdrawal and physical stop are equivalent responses. They are not. Withdrawing authorization from a robot mid-task — in particular, mid-transfer of a patient from bed to wheelchair — may itself create the harm the governance architecture exists to prevent [11, 12].

ARC-2 establishes that governance authority withdrawal and physical stop are categorically distinct governance actions, and that the decision to stop is itself a risk management decision governed by the same principles as the decision to continue. IEC 60204-1 defines three stop categories: Category 0 (immediate removal of power, uncontrolled), Category 1 (controlled deceleration before power removal), and Category 2 (controlled stop with power maintained) [9]. The choice among these is a risk assessment outcome, not a default.

ISO 13850 §4.1.1.3 explicitly provides that emergency stop "shall not be applied as a substitute for safeguarding measures" [10]. SAE J3016 §8.5 defines the Minimal Risk Condition as a state achievable through a Fallback DDT that considers both the risk of continuing and the risk of stopping [13]. EU AI Act Article 14(4)(e) requires a stop procedure that allows the system "to come to a halt in a safe state" — a safe-state halt, not unconditional immediate deactivation [14]. For personal care robots — the category most relevant to patient transfer scenarios — ISO 13482 provides the applicable safety standard [15].

ARC-2 specifies the following graduated safe transition protocol, applicable when any governance trigger crosses the threshold for scope contraction or higher:

(1) Fallback control activation: the system transitions to a reduced-capability operating mode with degraded but defined safety margins. Full LEA authorization is withdrawn; the system continues operating under a constrained fallback envelope.

(2) Reduced autonomy and velocity: operational speed and autonomy scope are reduced to the minimum required to complete the current task to a safe intermediate state.

(3) Task completion to safe intermediate state: the system completes the current action to a defined safe intermediate state (e.g., patient returned to supported position; load set down at nearest safe point) before any further authority reduction.

(4) Human operator handover: the system requests handover to a qualified human operator, specifying the current system state, the triggering condition, and the handover requirements. The LEA authorization must specify: the required qualifications of the receiving operator, the maximum latency for handover response, and the fallback procedure if no qualified operator is available within that window.

(5) Controlled stop: if handover cannot be completed within the specified latency window, the system executes a Category 1 or Category 2 controlled stop depending on current task state and environmental conditions.

(6) Emergency stop: Layer 1 hard safety invariant breaches (Section 5.1) may require immediate Category 0 stop regardless of task state.

Whether authority withdrawal constitutes a safety function requiring SIL/PL rating under IEC 62061 [16] or ISO 13849 [17] is an open question that ARC-2 acknowledges but does not resolve. This question must be addressed in the deployment-specific LEA authorization process.

### 5.3 Staged Response to Confirmed Spirals

ARC-2 proposes a staged response to confirmed Layer 3 spirals — not binary revocation:

Scope contraction: LEA authorization narrows to a lower-risk operational envelope.

Mandatory CRAB re-evaluation: triggered immediately upon spiral confirmation.

LEA suspension: if the spiral continues through re-evaluation, authorization is suspended pending investigation.

LEA revocation: if re-evaluation confirms the delta is not recoverable without model retraining or hardware intervention.

The staged approach preserves operational continuity during investigation and ensures that revocation is based on confirmed evidence rather than a single adverse reading.

## 6. Enhanced Performance: Asymmetric Governance

### 6.1 The Ambiguity of Positive Delta

A system outperforming its certified CRAB baseline presents an interpretive problem. The positive delta may reflect genuine improvement through continued exposure to operational environments. Or it may reflect metric gaming, distributional shift toward easier tasks, or degradation on unmonitored dimensions. Positive delta is not self-evidently good.

### 6.2 The Asymmetry Principle

ARC-2 proposes an intentional asymmetry: the governance burden for expanding LEA authorization is higher than the governance burden for contracting it. Contraction is triggered automatically by confirmed delta thresholds. Expansion requires human-confirmed review.

The rationale: restricting an authorized envelope when evidence of drift appears is conservative and recoverable. Expanding an authorized envelope on the basis of a metric improvement that may not generalize is neither. The asymmetry reflects the irreversibility of harm from over-authorization relative to the operational cost of delayed expansion.

### 6.3 The LEA Expansion Pathway

Sustained positive delta across all monitored dimensions, over a defined minimum window with no adverse events, opens an LEA expansion pathway — a formal eligibility for scope review. The pathway does not trigger expansion; it triggers eligibility for human-confirmed review. Expansion occurs only when a designated oversight body confirms that the improvement is genuine, generalizable, and consistent across all four delta dimensions.

## 7. Implementation Implications

### 7.1 Governance Infrastructure Requirements

Delta management requires granular, real-time telemetry. The compliance signal cannot be computed without continuous measurement of system state across all four delta dimensions. The telemetry stack must be treated as a first-class component of the governance architecture, not an afterthought.

The practical architecture of continuous monitoring depends on the proxy-versus-full-evaluation distinction. Proxy streaming must be low-latency and always-on; triggered full evaluation requires access to the certification test environment or its equivalent. These are architecturally different requirements that must not be conflated.

Model delta assessment must incorporate version hashes and update attestation records as necessary (though not sufficient) signals. For systems with continual learning or retrieval-augmented components, delta must be measured against a pre-approved change envelope rather than a static version record, consistent with the FDA Predetermined Change Control Plan framework [3].

### 7.2 Latency as a Governance Metric

The latency between a detected delta threshold breach and the corresponding governance response is itself a compliance metric. An architecture that detects drift but responds slowly provides only nominal protection. ARC-2 proposes that maximum permissible response latency be specified as part of the LEA authorization.

### 7.3 The Auditability Requirement

Delta management makes the "black box" approach to autonomy a governance liability. Mandatory logging of delta readings, governance events, and response actions across the full operational lifecycle is required. Logs must be tamper-evident (cryptographic hash-chaining or append-only signed structures), timestamp-anchored to a trusted external reference, and chain-of-custody documented. These requirements align with EU AI Act Articles 12 and 72 [6, 7].

For long-lived systems, storage volume and integrity requirements are non-trivial: a system generating delta readings at one-second intervals over a five-year deployment produces on the order of 150 million log entries per dimension before compression. The LEA authorization process should require that the operator's logging infrastructure meet defined standards before deployment approval is granted.

### 7.4 The Baseline Anchoring Problem

The delta management framework depends on the stability of B as a reference. B must include not only model weights and CRAB benchmark scores at the point of authorization, but also the environmental distribution under which certification testing was conducted. Without this specification, the distance function $d(S(t), B)$ is underspecified.

Environmental distribution shift is therefore a delta dimension in its own right — one with a distinct governance response as described in Section 3.3. When environmental drift exceeds a defined threshold, a governance event is triggered requiring review of whether the existing baseline remains an adequate compliance reference, or whether re-certification under the new environmental conditions is required.

### 7.5 Monitor Integrity

The monitoring layer is itself an attack surface and a single point of failure. Falsified delta signals can force dangerous stops through denial-of-service or mask genuine drift through suppression. The governance architecture must therefore require that the monitoring infrastructure satisfy integrity requirements including redundancy, mutual attestation between monitoring components, and anomaly detection on the monitoring data stream itself.

Applicable standards include IEC 62443 for industrial cybersecurity [18] and EU AI Act Article 15, which requires that high-risk AI systems be "resilient against attempts by unauthorised third parties to alter their use, outputs or performance by exploiting system vulnerabilities" [19]. These requirements apply to the monitoring layer with the same force as to the system under governance.

### 7.6 Recovery Pathway

ARC-2's staged response protocol addresses suspension and revocation in detail. What it must also specify is the pathway for recovery — the process by which a suspended or authority-contracted system may regain its authorized operational scope.

ARC-2 proposes a structured recovery pathway: (1) shadow mode operation, in which the system executes planned actions but all consequential outputs are human-confirmed before execution; (2) supervised trial period, in which the system operates with full human oversight across a defined duration and task set; (3) audit-gated restoration, in which a formal audit of the delta record, root-cause analysis of the triggering event, and remediation confirmation must be completed before re-authorization; and (4) revalidation, which for revoked systems requires full CRAB re-certification under current conditions before LEA authorization is considered.

A governance architecture that can only tighten authorization without defining a credible recovery pathway creates incentive structures that undermine compliance: operators who know that any suspension is permanent have incentives to avoid detection rather than to correct. The recovery pathway must be as clearly specified as the escalation pathway.

## 8. Discussion

### 8.1 Positioning Against Related Work

The contribution of ARC-2 must be understood in relation to several mature literatures that address adjacent problems.

Runtime assurance and the Simplex architecture [20] provide a well-established framework for switching between a high-performance controller and a provably safe backup when the former exits a verified safe envelope. ARC-2's delta management differs in that it is not primarily a switching architecture but a continuous authorization management framework: rather than switching controllers when safety is at risk, it continuously adjusts the authorized operational scope in response to measured deviation from the certified baseline. The two approaches are complementary rather than competing.

Drift detection methods [21] identify statistical changes in data distributions over time and are well-established for detecting covariate shift, concept drift, and model degradation. ARC-2's environmental delta dimension draws on this literature but repurposes it: the relevant question is not whether a distribution has shifted but whether the shift has taken the system outside its certified ODD in a way that renders the certified baseline an inadequate compliance reference. The governance coupling — that ODD exceedance triggers recertification rather than authority contraction — is the contribution, not the distributional shift detection itself.

Dynamic assurance cases [22] extend the safety cases literature to address post-deployment evolution, incorporating runtime evidence into formal safety arguments. ARC-2's delta management shares the ambition of making safety assurance continuous, but operates at the governance authorization level rather than the formal verification level. The frameworks are potentially complementary: ARC-2's delta readings could serve as runtime evidence inputs to dynamic assurance cases.

Post-market monitoring frameworks for medical AI [5] establish requirements for ongoing performance surveillance after regulatory approval. The ARC-2 framework is consistent with these requirements and may be interpreted as providing an operational specification for how post-market monitoring obligations are satisfied in autonomous robotic systems.

The central claim of ARC-2 — its defensible originality — is the coupling of continuously measured delta across a four-dimensional taxonomy to authorization state within a three-layer governance stack, including environmental delta as a baseline validity condition with a governance response distinct from the other three dimensions. This coupling has not, to the authors' knowledge, been formalized in the prior literature in this form.

CRAB-H [25] introduces task-centric certification for healthcare autonomous systems, establishing Task Requirement Profiles as governance objects that formally represent task cognitive requirements independently of system implementation. ARC-2's delta management framework is designed to generalize the CRAB-H approach: the certified baseline B encompasses not only the model safety and cognitive capability dimensions established at certification, but also the task-specific requirements defined in any applicable TRP-H. Runtime monitoring evaluates whether demonstrated capabilities continue to satisfy task requirements throughout deployment — not only whether behavior has drifted from an undifferentiated baseline.

### 8.2 Positioning Against EU AI Act Article 72

"Continuous compliance" is not only the conceptual framing of ARC-2 — it is existing regulatory terminology. EU AI Act Article 72(2) requires that providers of high-risk AI systems establish post-market monitoring systems to "evaluate the continuous compliance" of those systems throughout their lifecycle [7]. Articles 9 and 17 establish the risk management and quality management obligations within which post-market monitoring operates [23].

ARC-2 is best understood as proposing an operational specification for how autonomous robotic system operators may satisfy these Article 72 obligations. The delta management framework provides the measurement architecture (what to monitor), the trigger taxonomy (when to respond), the safe transition protocol (how to respond to physical systems), and the auditability requirements (how to document compliance) that give Article 72's requirements operational content in the autonomous robotics context.

The explicit positioning strengthens the paper's claim to practical relevance: rather than proposing an additional governance layer, ARC-2 offers an implementation architecture for requirements operators already face. Following the Digital Omnibus on AI [24], the application dates for certain high-risk obligations have been deferred; this does not diminish the relevance of the framework, as the underlying obligations remain.

### 8.3 Open Empirical Questions and Invitation for Contributions

ARC-2 is a governance architecture proposal, not an empirically validated system. Several of its core claims depend on empirical questions the authors cannot resolve from first principles alone. We identify these openly and invite contributions from the research and practitioner communities.

Trigger calibration. The layered trigger taxonomy in Section 5.1 requires domain-specific calibration of threshold values for each layer. The authors have no operational deployment data for specific robotic systems and therefore cannot propose validated thresholds. We invite researchers and operators with access to deployment telemetry to contribute threshold estimates, false positive and false negative rates, and Average Run Length observations across the four trigger layers. The framework is designed to accommodate such contributions without architectural modification — threshold values are deployment-specific parameters within the LEA authorization, not fixed constants.

Outlier floor quarantine performance. The quarantine mechanism proposed in Section 4.2 balances false floor formation against delayed response to genuine failures. The optimal quarantine window and confirmation criteria depend on system type, measurement cadence, and the statistical properties of sensor noise in specific deployment environments. We invite empirical characterization of these parameters, particularly in healthcare and logistics robotics where measurement conditions are well-documented.

Environmental delta measurement. Section 3.3 proposes monitoring distributional distance between operational sensor data and the certification environment, but does not specify which distributional distance measures are most appropriate for which sensor modalities, or what threshold values correspond to governance-relevant ODD exceedance. These are empirical questions. We invite contributions from researchers working on distribution shift detection, domain adaptation, and ODD monitoring in operational robotics.

Safe transition protocol timing. The graduated protocol in Section 5.2 specifies a sequence of steps but does not quantify the time budgets associated with each step. The maximum tolerable latency between fallback activation and safe intermediate state completion is task-specific and cannot be determined without task analysis data. We invite practitioners to contribute timing analyses for specific healthcare and logistics tasks, which could form the basis for domain-specific LEA authorization templates.

Researchers or practitioners with relevant data, simulation environments, or deployment experience who wish to contribute to the empirical foundation of this framework are encouraged to contact the corresponding author. The governance architecture is designed as a living framework: empirical contributions that refine or challenge its components are explicitly welcomed.

### 8.4 Limitations

ARC-2 has several limitations that must be explicitly acknowledged.

(a) Absence of empirical validation. The paper presents a governance architecture without empirical demonstration that the delta management framework, applied to real or simulated operational traces, produces the detection characteristics claimed. No receiver operating characteristic curves, false positive or false negative rates, or simulation back-tests are provided. The authors regard this as the most important open gap in the framework. We explicitly invite contributions from researchers and practitioners who have access to operational deployment data, simulation environments, or robot platform telemetry. Specifically, we welcome: (i) empirical evaluations of trigger layer performance across the four-tier taxonomy using real or synthetic traces, including characterization of $ARL_0$ and positive predictive value at realistic base rates; (ii) case studies documenting environmental delta threshold crossings and the governance responses they triggered; (iii) operational data on CRAB proxy monitoring reliability and calibration drift; and (iv) comparative assessments of the outlier floor quarantine mechanism against alternative robustness strategies. Authors interested in collaboration or data sharing are encouraged to contact the corresponding author.

(b) Underspecified distance metric. The compliance signal $\Delta(t)$ is specified as a distance-to-set function without fully specifying the normalization of heterogeneous dimensions, the handling of missing data, the temporal aggregation method, or the uncertainty bounds. A complete implementation specification requires these choices to be made and documented as part of the LEA authorization. The authors invite practitioners with experience in multivariate process monitoring, statistical process control, or runtime verification to propose and evaluate concrete distance metric specifications for specific deployment categories.

(c) SIL/PL classification of authority withdrawal. Whether the authority withdrawal mechanism constitutes a safety function requiring Safety Integrity Level (SIL) or Performance Level (PL) rating under IEC 62061 [16] or ISO 13849 [17] is an open question. The answer depends on deployment context and the failure modes of the withdrawal mechanism itself. This paper acknowledges the question but does not resolve it.

(d) Partial regulatory role mapping. EU AI Act Articles 16 and 26(5) establish distinct obligations for providers and deployers of high-risk AI systems. The distribution of ARC-2's governance obligations between these roles — in particular, whether LEA authorization is a provider obligation, a deployer obligation, or a shared obligation — is not fully specified. Future work should address this mapping explicitly.

## 9. Conclusion

ARC established that autonomous robotic systems require a three-layer governance architecture spanning model safety, cognitive certification, and operational authorization. This paper has argued that the architecture must be animated by a governing principle for post-deployment compliance: delta management.

The paper reframes operational compliance as a continuously measurable property rather than a binary certification outcome. Certification establishes the reference state — a certified region rather than a point; delta management determines whether that reference state is preserved throughout deployment, across four dimensions: model behavior, cognitive capability, operational performance, and environmental validity.

The four extensions proposed — outlier-indexed CRAB scoring with quarantine and rebaseline pathways, a layered trigger taxonomy that distinguishes hard invariants from single-dimension overrides and confirmed spirals, a graduated safe transition protocol that separates authority withdrawal from physical stop, and environmental delta as a baseline validity condition with a distinct recertification response — follow directly from the delta management principle and build naturally on the ARC-1 framework without requiring new foundational architecture.

The practical consequence is a shift in what "compliant" means. A compliant system is not one that passed certification. It is one that, at every moment of its operational life, can demonstrate that it remains within the authorized distance from its certified baseline — and that the baseline itself remains a valid reference for the environment in which it operates.

## Competing Interests

The authors declare the following competing interests: Wybe Labs Inc. is a commercial entity developing governance tools for autonomous systems. This paper does not constitute a regulatory standard. Compliance with any standard requires independent conformity assessment.

## Disclaimer

This paper does not constitute an adopted standard or regulatory guidance. References to certification and authorization describe a proposed governance architecture, not a claim of regulatory compliance under any jurisdiction. Nothing in this paper should be construed as legal advice or as a representation that compliance with the proposed framework satisfies any applicable regulatory requirement.

## References

References are numbered in order of first appearance in the text.